\documentclass{article}

\usepackage[preprint]{neurips_2023}

\usepackage[utf8]{inputenc} % allow utf-8 input
\usepackage[T1]{fontenc}    % use 8-bit T1 fonts
\usepackage{url}            % simple URL typesetting
\usepackage{booktabs}       % professional-quality tables
\usepackage{amsfonts}       % blackboard math symbols
\usepackage{nicefrac}       % compact symbols for 1/2, etc.
\usepackage{microtype}      % microtypography

\usepackage{graphicx}
\usepackage{amsmath}
\usepackage{amssymb}
\usepackage{booktabs}
\usepackage{bm}
\usepackage{caption}
\usepackage{float}
\usepackage{blindtext}
\usepackage{array}
\usepackage{tabularx}
\usepackage{booktabs}
\usepackage{ragged2e}
\usepackage{color, colortbl}
\usepackage{makecell}
\usepackage{caption}
\usepackage{bbding}
\usepackage{authblk}

\definecolor{Gray}{gray}{0.9}
\definecolor{LightCyan}{rgb}{0.88,1,1}

\usepackage{multirow}
\usepackage[normalem]{ulem}
\useunder{\uline}{\ul}{}
\usepackage[breaklinks,colorlinks]{hyperref}

\usepackage[capitalize]{cleveref}
\crefname{section}{Sec.}{Secs.}
\Crefname{section}{Section}{Sections}
\Crefname{table}{Table}{Tables}
\crefname{table}{Tab.}{Tabs.}

\newcommand*\samethanks[1][\value{footnote}]{\footnotemark[#1]}

\title{Label Name is Mantra: \\Unifying Point Cloud Segmentation across Heterogeneous Datasets}

\author{Yixun Liang\thanks{Equal contribution.} }
\author{Hao He\samethanks}
\author{Shishi Xiao}
\author{Hao Lu}
\author{Yingcong Chen\thanks{Corresponding author}}
\affil{Hong Kong University of Science and Technology (Guangzhou)}

\begin{document}

\maketitle

\begin{center}
    \centering
    \captionsetup{type=figure}
    \includegraphics[width=\textwidth]{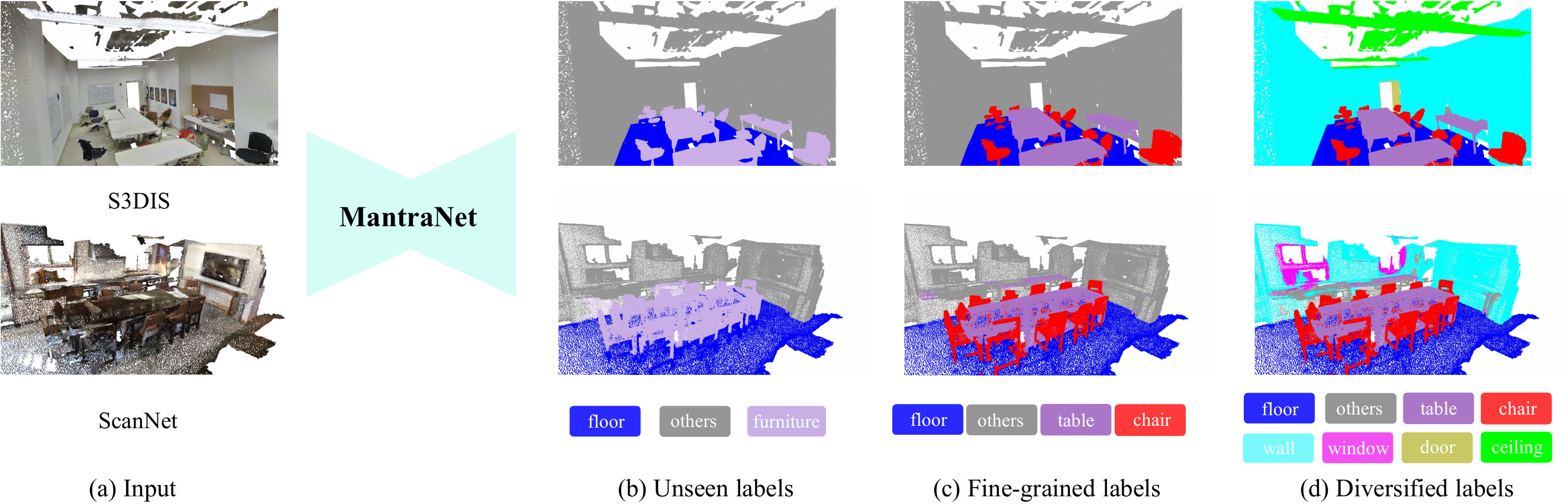}
    \captionof{figure}{The proposed unified framework MantraNet can handle point clouds from multiple datasets with different labels, and it shows robust generalization ability in unseen labels, fined-grained labels and diversified labels as shown in (b)(c)(d).
    } 
    \label{fig:teaser}
\end{center}

\begin{abstract}
Point cloud segmentation is a fundamental task in 3D vision that serves a wide range of applications. Although great progresses have been made these years, its practical usability is still limited by the availability of training data. Existing approaches cannot make full use of multiple datasets on hand due to the label mismatch among different datasets. In this paper, we propose a principled approach that supports learning from heterogeneous datasets with different label sets. Our idea is to utilize a pre-trained language model to embed discrete labels to a continuous latent space with the help of their label names. 
This unifies all labels of different datasets, so that joint training is doable. Meanwhile, classifying points in the continuous 3D space by their vocabulary tokens significantly increase the generalization ability of the model in comparison with existing approaches that have fixed decoder architecture.
Besides, we also integrate prompt learning in our framework to alleviate data shifts among different data sources. Extensive experiments demonstrate that our model outperforms the state-of-the-art by a large margin.  
\end{abstract}

\section{Introduction}
% Semantic segmentation in point cloud is a fundamental and challenge problem for 3D scene understanding, with the aim of classifying every 3d coordinates with their respective semantic labels. Recently, it has drawn a wide range of attention since it is a major building block for autonomous driving, VR/AR, and robotics. 

Perceiving and understanding the 3D world is one of the most important tasks in computer vision. The rapid development of 3D sensors, e.g., Lidar or Kinect, makes 3D information much more accessible than ever. Among them, point cloud is one of the most important modalities obtained directly from 3D sensors, and point cloud segmentation is indispensable in a wide range of downstream applications, such as autonomous driving, VR/AR, and robotics. 

% Although previous methods \citep{qi2017pointnet,thomas2019kpconv,qian2022pointnext} made significant advances to improve the performance of 3D semantic segmentation system, there are two intrinsic limitations have been ignored:\emph{data scarcity and close-set}. Data driven approaches show great performance in 3D semantic segmentation. However, collecting large amount of 3D point cloud data is expensive and time-consuming. Thus, many existing methods have compromised to use fixed label sets and align hierarchy labels, which cause the network vulnerable to data distribution shift. The close-set network assumes that all categories are known during training. If a novel class is encountered during inference, it will still assign the old labels to the novel class, limits its usage for real world applications.

In recent years, point cloud segmentation has drawn increasing attention. Existing works \citep{qi2017pointnet,thomas2019kpconv,qian2021assanet,zhao2021point,qian2022pointnext,lai2022stratified} mainly focus on developing new architectures to improve accuracy and efficiency. Despite the great progress, these works may face difficulties in real-world applications. The reasons are manifolds. Firstly, as deep learning-based methods, these models can not be well optimized without a sufficient amount of samples. In addition, labeling point clouds is very expensive, and making such methods hard to migrate to real-world applications. Besides, these models can only make predictions on the training label set while showing inferior performance on the unseen domain. However, due to the ever-changing practical needs, one may need to make predictions beyond training labels. As shown in Fig.~\ref{fig:teaser}(b)(c), Some unnoticed objects may be requested next. In this scenario, users have to re-collect and re-label data, which costs a lot of effort. 

%  A straightforward method is to leverage all available datasets and train one unified mode, which increase the amount of training data and diversity to prevent overfitting. Moreover, introducing more labels improves model's generalization ability, and making it less error prone to novel classes. However, directly combining all different datasets for training is not trivial with current paradigms \textbf{\{citation\}}. On one hand, current paradigms with fixed-size decoder consider each category as parallel and directly regress the discrete probability of such categories, which overlook the fact that labels from different datasets have different level of granularity. On the other hand, category agnostic shift \textbf{\{citation\}} should be considered when training with heterogeneous data sources, as we can no longer assume domain shift is negligible.  

\begin{figure}[!t]
 \centering
   \includegraphics{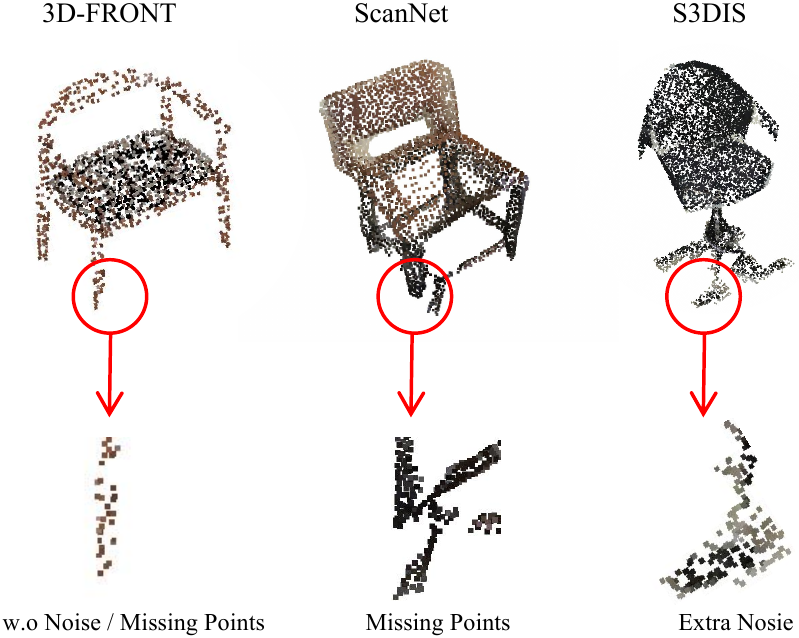}
\caption{\textbf{An example of data shift.} Although those three instances from heterogeneous sources shown above are all annotated as "\textit{chair}", there are several differences caused by external reasons. On the one hand, chairs for diverse applications differ greatly in appearance. On the other hand, since the point clouds are manually scanned, there are no missing points and extra noises in the simulated source (3D-FRONT) compared to the real sources (ScanNet, S3DIS).}
    \label{fig:ca_shift}
\end{figure}

To deal with these problems, one straightforward idea is to leverage all datasets on hand to train a unified model. This on the one hand, increases the number of training data to prevent overfitting; on the other hand, it also enlarges the label set to meet a wider range of practical needs. However, directly combining all datasets for training is non-trivial in practice. As different datasets are labeled differently, joint training requires merging all labels together. 
% However, these labels are not independent with each other \yc{refer to a figure}, and thus directly merging them leads to severe confliction.
In addition, there exists a lot of dependencies among the labels, such as synonyms or labels with different granularity, and thus directly merging them leads to severe conflict.
Another option is to manually align labels of different datasets~\citep{ding2022doda}, i.e., merge synonymous labels to one label. However, this is not always achievable, as labels of different datasets may have different granularity. Like Fig.~\ref{fig:teaser}(c)(d), "\textit{table}" and "\textit{chair}" can be categorized as part of the "\textit{furniture}". Besides the aforementioned problem, there also exists covariate shift as data from different sources are collected in different ways, see Fig. \ref{fig:ca_shift} for example. 
% \yc{TODO: category agnostic shift}

% In this paper, we propose \textbf{\{Name\}}, a principled framework that allows training with heterogeneous data sources under open-set assumption for 3D semantic segmentation. 
To overcome these issues, we propose the MantraNet, a principled approach that allows unified training with heterogeneous data sources. Our framework supports training from multiple datasets, regardless of the fact that their label sets may be different. 
Besides, during testing, our model works well with unseen labels without re-training. 
Our key idea is that, instead of treating labels as one-hot vectors as usual practice, we fully leverage label names to model the intrinsic relation between different classes. 
% Specifically, we propose the semantic label encoding that embeds label names in a continuous latent space where different categories and granularity can be well represented.
This is achieved by encoding label names to a continuous latent space with a pre-trained language model, where labels of different categories and granularity can be well represented. As such, similar labels in different datasets are clustered together, while dissimilar ones are located far away.  
In this sense, during training, we can align point cloud features with the encoded label names for all datasets. This makes the feature space maintain a well-defined semantic topology, so that it works well with misaligned labels in both training and testing. 
Furthermore, we also propose a novel module to deal with data shifts across different datasets in our model. Motivated by~\citep{ge2022domain,zhou2022conditional,zhou2022learning}, we concatenate label name tokens with domain-specific prompts, so as to encourage learning a category and domain disentangled representation. Different from image recognition tasks that prompt can be manually designed, in point cloud segmentation, prompt engineering is very challenging. Therefore, we propose to learn these prompts in an end-to-end manner. 
To summarize, our contributions are listed as follows. 
\begin{itemize}
    \item We present a principled framework that supports training with multiple datasets. Our approach can automatically align different label sets in both the training and inference stages. 
    \item We incorporate a novel prompt learning module in our framework, which alleviates covariate shifts across different datasets. 
    \item We conduct extensive experiments on a large benchmark to validate the effectiveness of our approach. We show that our method outperforms the state-of-the-art by a large margin. 
\end{itemize}

% In summary, our contributions are 3 folds: 

% 1) A Novel framework called \textbf{\{Name\}} that leverages a pretrained language model with 3D point cloud representation, which adds continuous semantic awareness into the 3D scene space.  
% %a framework leverage 3d point cloud with clip, that adds continous semantic awareness? (update later_)

% 2) A light weight prompt module that alleviate domain shift when training multiple benchmarks together.   
% %a novel prompt module that alleivate domain fluctuation when train mulitple benchmarks together (something like that)

% 3) Based on our best knowledge, we are the first to define a open-set framework that jointly optimizing CLIP with 3D point cloud semantic segmentation. Experiment results show that our proposed framework is superior to other baselines.  
% %turn close-set issue to open-set (treat latent space differently)
% %%%%%%%%%%%%%%%%%   Related Work    %%%%%%%%%%%%%%%%%%%%%%%%%%
\section{Related Works}
\label{sec:related}
\paragraph{Point cloud semantic segmentation}
\label{ssec:seg}
% Given an unstructured point cloud, semantic segmentation aims to predict semantic labels for each 3D point.
Point cloud segmentation aims to predict semantic labels for each 3D point. 
Existing methods can be divided into two categories, i.e., voxel-based and point-based.
Voxel-based approaches~\citep{maturana2015voxnet, choy20194d,graham20183d, graham2017submanifold} consider points as regular voxels and transform them into a grid-like structure, which could be processed by standard 3D convolutional neural networks. 
% Therefore, many existing approaches leverage such structure with 3D convolutional neural networks on point cloud segmentation and achieve decent performance. 
% However, voxelization must consider the trade-off between precise geometry information and memory cost.
However, these approaches require a voxelization process, and thus there exists a clear trade-off between precision and memory cost. 
In contrast, point-based approaches directly operate on each point, which maintains accurate scene geometry with affordable memory cost.
PointNet~\citep{qi2017pointnet} first uses point-wise MLPs and pooling layers to achieve permutation-invariant constraint and symmetric aggregation.  Following its success, various techniques are proposed to enhance feature extraction and aggregation structures. Among them graph-based techniques~\citep{landrieu2018large, wang2019dynamic} aims to capture semantic characteristics, 3D continuous convolution techniques~\citep{thomas2019kpconv, wu2019pointconv, boulch2020fkaconv,xu2021paconv} propose to enhance flexibility and efficiency, and transformer-based techniques~\citep{zhao2021point,guo2021pct,lai2022stratified} leverage attention mechanism, which aggregates local and long-range contextual features.

However, previous works mostly focused on improving the performance of point cloud semantic segmentation tasks, with an assumption that all labels are annotated and task objectives are fixed. % This assumption ignores the fact that practical needs are changing frequently. 
When practical needs changes, these approaches need to re-collect or re-label a lot of data, which is costly. Besides, these works do not support joint training with multiple heterogeneous  datasets where data distribution and label sets are different. 
% Our works resolve this issue by joitly considering points' semantic representation and appearance, and are able to adapt to the changing needs.
Our work provides a principled solution to  resolve these problems by mapping labels to a shared semantic space with their names. In this sense, it can significantly improve the usability in real world scenarios. 

\paragraph{Text-guided learning}
\label{ssec:VLP}
Our framework relies on pre-trained language models. 
Language models have shown promising performance in both natural language processing and vision tasks these years.
One common approach is to leverage text embedding as prior knowledge to facilitate the process of learning visual representation~\citep{li2019visualbert, ramesh2021zero,tan2019lxmert}.
Such a method typically uses Faster-RCNN~\citep{ren2015faster} as a feature extractor to find the region of interest (ROI), and jointly consider its appearance and semantic representation.
SimVLM~\citep{wang2021simvlm} reduces the need of using a feature extractor by using a large-scale weakly self-supervised training with a prefix language model~\citep{raffel2020exploring}.
With a high-capacity text encoder and visual encoder, CLIP~\citep{radford2021learning} aligns of text and visual elements in heterogeneous embedding space.
Several studies extend the applicability of CLIP to other tasks such as object detection~\citep{gu2021zeroshot}, semantic segmentation~\citep{wang2022cris, li2022language, zhang2022pointclip} and image editing~\citep{patashnik2021styleclip,gal2021stylegan}. 
Among them, PointClip~\citep{zhang2022pointclip} is most related to our work. It learns transferable visual concepts by leveraging CLIP’s pre-trained knowledge for zero-shot point cloud recognition.
% However, it requires converting 3D data to multi-view depth maps that contain geometry information, hindering the performance of visual recognition. % 有信息损失
% Moreover, this framework lacks of the ability to perform low-level segmentation tasks since it requires processing information at each point directly.
However, it needs to project the 3D data to 2D depth maps, which inevitably losses information. Besides, this also prevents it to be extended to dense prediction tasks like point cloud segmentation. 
Our model is advantageous in the sense that we directly work on 3D point clouds, and thus can  preserve all information and be applicable for the dense point cloud segmentation task. 
% In contrast, our method provides a novel way to utilize CLIP or other pre-trained language model's knowledge by jointly considering the appearance and semantic representation of point cloud data. 
% In contrast, our method provides a novel way to utilize CLIP or other pre-trained language model's knowledge by utilizing label names 
% It broadens the probability of solving low-level recognition tasks by distilling the pre-trained language model's knowledge.

\paragraph{Prompt learning}
\label{ssec:Pl} 
Our framework also incorporates a prompt learning module. 
In contrast to the transfer learning schema that directly finetunes on downstream tasks, prompt learning proposes a novel paradigm for aligning downstream tasks to the pre-trained model.
The emergence of large vision-language model, \emph{e.g.}, CLIP~\citep{radford2021learning} and ALIGN~\citep{jia2021scaling}, promotes zero-shot and generalization capacity significantly in several visual tasks~\citep{wang2022cris, ju2021prompting, li2022language}.
Recently, many works~\citep{zhou2022learning, zhou2022conditional,khattak2022maple,lu2022prompt} explored learnable prompts for retrieving more accurate representation contained in the pre-trained model.
CoOp~\citep{zhou2022learning} surpasses the performance of CLIP by context optimization at the language branch, and  CoCoOp~\citep{zhou2022conditional} extends the generalization ability by generating a conditional token for image instances.
As using a single branch is sub-optimal for flexibly adapting to multi-modal representation space, MaPLe~\citep{khattak2022maple} leverages both language and image branches to enhance alignment with coupling function.
Additionally, some works~\citep{zheng2022prompt,ge2022domain} embed the source domain knowledge into prompts for downstream tasks in the target domain.
In this work, we explored the application of prompt learning in point cloud segmentation. Intending to separate category-independent scene representations via statistical properties of unordered point features, we leverage the knowledge of large pre-trained models to improve generalization.

\paragraph{Label embedding} 
Label embedding ~\citep{akata2015label,frome2013devise,socher2013zero,palatucci2009zero,li2015semi,sun2017label} encodes vanilla one-hot vector labels to a latent space with semantic topology, which is widely used to solve the zero-shot problem.
Our approach is related in that we encode the labels to the semantic language space by the label names. 
% This approach is widely used in computer vision~\citep{akata2015label,frome2013devise,palatucci2009zero,socher2013zero} to make the model more explainable and solve the zero-shot problem.
% However, As label embeddings only learned from training data, these methods suffer from overfitting to training data. Meanwhile, xxx and xxx try to leverage pre-trained model to make label embeddings be more accurate.
% Previous works generate label embedding manually~\citep{palatucci2009zero} or by deriving the semantic relationship between labels with language model~\citep{frome2013devise}.
\citep{palatucci2009zero} encodes each class with semantic features that characterize a wide range of possible classes, allowing the model to achieve zero-shot capacity by extrapolating the relationship between unseen input and semantic knowledge.
With outlier detection in semantic space, \citep{socher2013zero} recognizes unseen classes and classifies them based on isometric Gaussian distribution.
As these methods simply construct the semantic space from unsupervised large text corpora, \citep{li2015semi,sun2017label} produce more adaptive embedding without side information by directly obtaining label embedding from input data.
Nevertheless, these works are limited by fixed label encoding, in which a set of samples can only be represented by one label.
Instead of following such a one-to-one representation, we extend label embedding by incorporating the alternative representation under the guidance of pre-trained language models. 
Moreover, by incorporating prompt learning, our labels are adaptive to different scenes, which leads to better generalization capacity to various domains. 
% Therefore, more informative features brought by several label embeddings enhance the performance of semantic segmentation.

%%%%%%%%%%%%%%%%%   Method    %%%%%%%%%%%%%%%%%%%%%%%%%%
\section{Methodology}
\begin{figure*}[!htbp]
 \centering
   \includegraphics[width=0.85\linewidth]{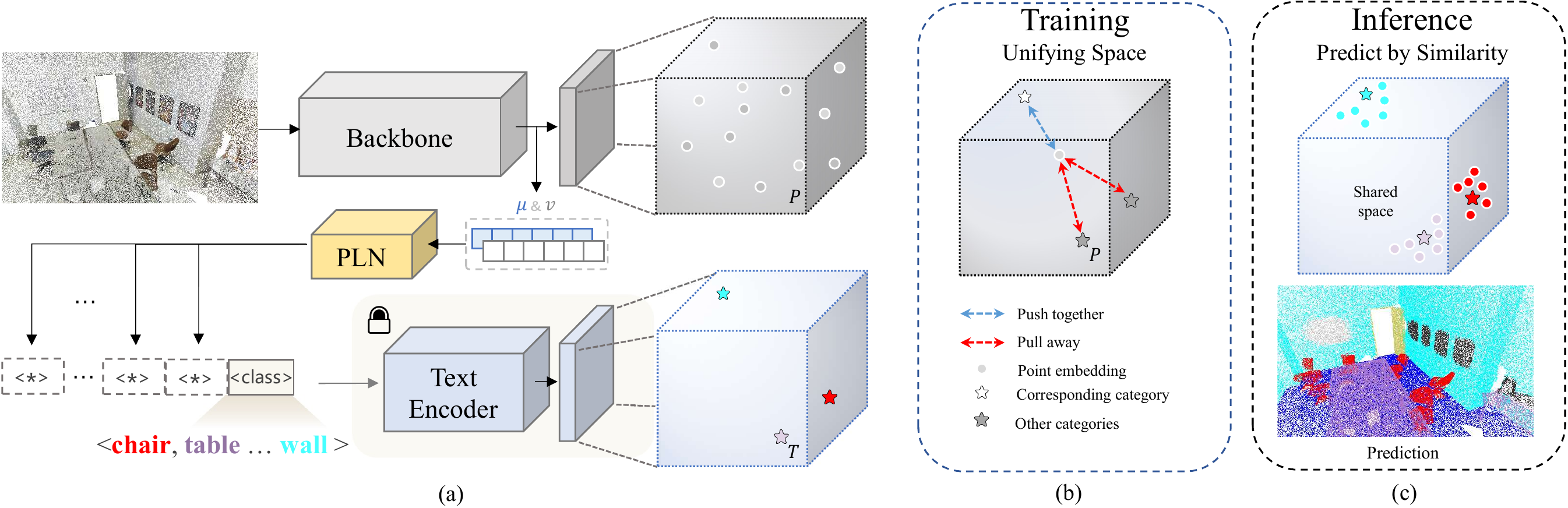}
\caption{(a) Overview of our model. To bridge point clouds and latent space provided by the text encoder, MantraNet first extracts pointwise features and project them to the same dimension as latent space. Then a Prompt Learning Network (PLN) conducts extracting scene-specific descriptions by the global representation of pointwise features. Such descriptions will be combined with label names as prompts and encoded by a text encoder. (b) Illustration of the training process. During training, we directly supervise space unifying by pushing the point embeddings to their corresponding label name's embedding and pulling away from others' label embeddings. (c) Illustration of the inference process. During inference, categories of each point will be classify via the similarity with given label embeddings.}
\label{fig:model}
\end{figure*}
% semantic awareness in countious space -> jointly considering vision and language representation -> solve fix decoder issue
% adapot to more benchmarks ease data scarcity issue -> found domain shift damage performance -> introduce light weight prompt module

\subsection{Overview}
In this section, we first formalize the definition of our problem, given $N$ training sets, the label names of each training sets denoted as $C_0= \left \{0,1,\cdots,n_0\right \}$, $C_1= \left \{0,1,\cdots,n_1\right \}$, $\cdots$, $C_N= \left \{0,1,\cdots,n_N\right \}$, unlike traditional methods consider them as separated tasks and train or inference them individually, we unify all of $N$ training sets to refine a generalized framework which can be trained or inferenced on any given labels $c_g$ that relate to the unified training set.

Instead of previous efforts setting labels as discrete one-hot vectors and directly regress possibilities of each predefined category by a fixed decoder, we utilize a pre-train language model to encode label names from all training sets $ C_{all} = C_0 \cup C_1...\cup C_N$ as semantic anchors. By leveraging such semantic anchors, we optimize the appearance representation of 3D point clouds to align with their corresponding label representation. During inference, our MantraNet classifies each point by the similarities between point and label embedding. The framework we proposed is shown in Fig.~\ref{fig:model} and we will explain it in detail in the following sections.

\subsection{Unified Supervision by Label Names}\label{Shared Space}
Given the unified training set $ C_{all} $, we first project label names to a semantic latent space by leveraging the pre-train language model denote as $F_{text}\left( \cdot \right)$, 
\begin{equation}
	\label{eq1}
	{\textbf T}_{k} ={F}_{text}\left ( C_{all} \right ),
\end{equation}
where $\textbf T_{1}, \textbf T_{2}, \cdots, \textbf T_{n_0+n_1+...n_N} \in \mathbb{R}^{D}$ are embeddings of corresponding classes in the semantic latent space of pre-trained language model. Label embeddings can be considered as anchors to obtain a network that maps 3D point clouds to the semantic latent space. Therefore, we are able to train heterogeneous datasets together because of the semantic anchors, rather than manually align labels by prior knowledge and naively optimize them as a single dataset.

% Pre-trained languange model has the semantic awareness of text informations, and it can handle labels even in heterogeneous datasets. Label embeddings can be considered as anchors to obtain a network that map 3D point clouds to the semantic latent space. Therefore, we are able to jointly train heterogeneous datasets together, rather than manually align labels by prior knowledge and train those datasets individually.
%语言模型可以很
%Pre-trained languange model has the semantic awareness of different kinds of labels, and it can handle labels even in heterogeneous datasets, as shown in Fig. N. Therefore, we are able to jointly train(optimize) heterogenous dataset together, rather than manually align labels by prior knowledge and train those datasets individually. Label embeddings can be considered as anchors to obtain a network that map 3D point clouds to the semantic latent space.
%Such label semantic represnetation % and that of 3D points.

% We use a feature projector to connect 3D point clouds with their labels' semantic representation by 
% To obtain a network that connects 3D point clouds with their labels' semantic representation, we first accomplish multimodal connectivity in the embedding level via feature projection.
Similar to traditional 3D semantic segmentation framework, we also leverage a point-based backbone to extract per-point features $ \hat{\textbf{p}}  \in \mathbb{R}^{N\times d}$ from the input scene $ \textbf p \in \mathbb{R}^{N\times6}$, and those features will be projected to a latent space with their label embeddings $\textbf T \in \mathbb{R}^{D}$. We use a linear layer as projector which denote as ${F}_{proj.}\left( \cdot \right)$, 
\begin{equation}
	\label{eq2}
	\textbf{P} ={F}_{proj.}\left( \hat{\textbf p} \right),
\end{equation}
\noindent
where $\textbf{P}\in \mathbb{R}^{N \times D}$ are corresponding point embeddings of the input scene $ \textbf p \in \mathbb{R}^{N\times6}$ and the embedding of point $(i)$ is refer as $\textbf{P}_{i}\in \mathbb{R}^{D}$. %We then implement linkage of points with semantic latent space.
Given the embedding $\textbf{P}_{i}$, we calculate the similarity between $\textbf{P}_{i}$ and label embeddings $\textbf T_{j},j \in \left \{1,2, \cdots, n_k \right \}$ from the corresponding label set $C_k$. Following CLIP~\citep{radford2021learning}, we choose cosine similarity $_{\!}$ $\langle\cdot,\cdot\rangle$$_{\!}$ as our measurement,
\begin{equation}
	\label{eq3}
	Sim_i=\langle \textbf P_{i}, \textbf {T}_{j} \rangle\!=\frac{\textbf P_{i}}{\left \|  \textbf P_{i}\right \|_{2}} \cdot \frac{\textbf T_{j}}{\left \|  \textbf T_{j}\right \|_{2}},j \in \left \{1,2,\cdots, C \right \},
\end{equation}
where $\left \|\cdot\right \|_{2}$ denote the L2-norm operator and the similarity between point embedding $\textbf{P}_{i}\in \mathbb{R}^{D}$ and given label embeddings are refer as $Sim_i \in \mathbb{R}^{C}$. Therefore, we define the probability distribution of this point over $n_k$ categories as,
\begin{equation} %\small
\begin{aligned}\label{eq:np}
\!\!\!\!\!p(c|\textbf P_{i})\!=\!\frac{\exp(Sim_{i,c}/t)}{\sum^{n_k}_{c'=1\!}\exp(Sim_{i,c'}/t)},
\end{aligned}
\end{equation}
where $t$ is the temperature parameter, we set $t=0.1$ following \citep{he2020momentum}. Therefore the cross-entropy loss with temperature scaling can be used to directly supervise the space alignment.
\begin{equation}\small
\begin{aligned}\label{eq:nce}
\mathcal{L}_{\text{CE}} &= -\log p(c_j|\textbf P_{i}) \\[-2pt]
&=  -\log \frac{\exp(Sim_{i,c_j}/t)}{\exp(Sim_{i,c_j'}/t)\!+\!\sum_{c'\!\neq c_j\!}\exp(Sim_{i,c_j'}/t)},\!\!
\end{aligned}
\end{equation}
%%%你看看怎么写更好
where the $c_j\in \left \{1,2,\cdots, n_k \right \}$ means ground-turth category of each points $(i)$. Intuitively, Eq.~\eqref{eq:nce} pull embedding of point $\bm i$ away from unrelated classes and push it to its corresponding one, which directly supervised the latent spaces alignment via the given anchors. 

\begin{figure}%[H]
\centering
\includegraphics[width=0.5\linewidth]{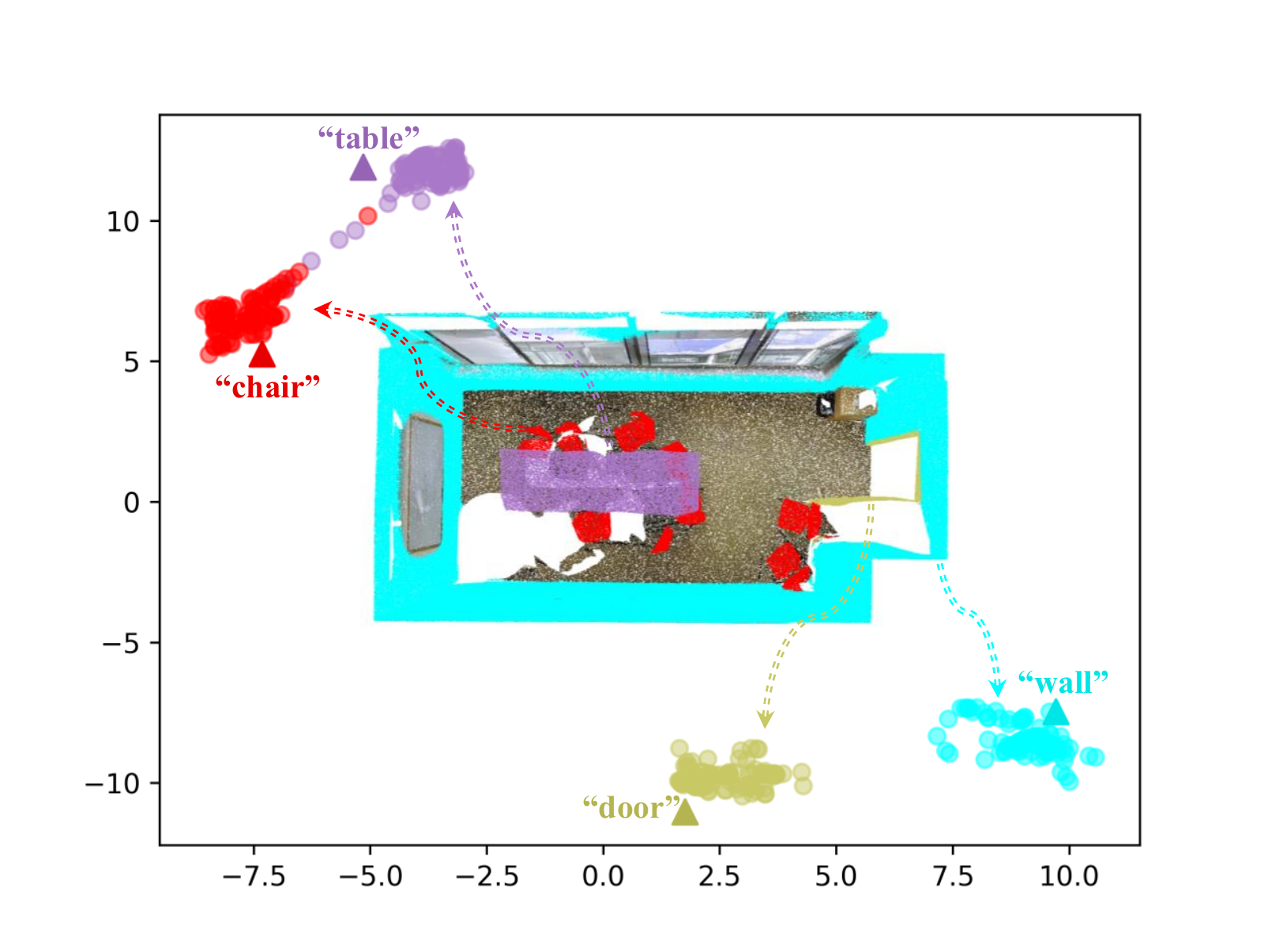} %0.5
\caption{\textbf{t-SNE visualization of object and text embeddings with our proposed MantraNet.} Point embeddings are represented as cycles and label embeddings are represented as triangles. the location of related objects and labels are closer compared to those that are not related in the shared space, which proves that semantic concepts can be modeled by distance naturally in the latent space.}
\label{tsne}
\end{figure}

In contrast to previous methods, our advantages are two folds. Firstly, the locations of objects in the semantic latent space correspond to their semantic distribution in the real world, which implicitly encourages network to extract category shared features and prevent performance degradation, as shown in Fig.~\ref{tsne}. Secondly, the pre-trained language model can encode a wide range of vocabulary, which enables zero-shot ability to our MantraNet on unseen labels via querying the similarity between point and given word's embedding. Since our method does not require pre-defined label sets, we remove the constraint that previous methods must have fixed decoder architecture. Thus, our method greatly increases generalization of 3D point cloud semantic segmentation. 
% This greatly increases the generalization of our MantraNet, the number and granularity of objects of interest no longer need to be fixed and predefined.

\subsection{Scene-Specific Prompt Learning}
We lift the limitations of previous works that have to train on heterogeneous datasets individually and improve the model generalization ability by using the method introduced in Sec.~\ref{Shared Space}. However, heterogeneous datasets often come with the issues of covariate shift, and the objects with the same label might have different appearances when they come from different datasets, as shown in Fig.~\ref{fig:ca_shift}. Such issues impact overall segmentation performance of the model. To alleviate such issues, we leverage prompt learning~\citep{gal2022image,zhou2022conditional,zhou2022learning}, which shows that learnable prompt can effectively summarize the semantic representation of images. Specifically, we add a Prompt Learning Network (PLN), which is a light weight network that decouples category-irrelevant information via points' high-level global features and encode them as prompt tokens. These tokens are scene-specific descriptions for labels to obtain a more accurate anchor in the latent space.
To be specific, given the extracted features of scenes $ \hat{\textbf{p}} \in \mathbb{R}^{N\times d}$, we first acquire global features via computing a per-element mean $\bm{\mu} \in \mathbb{R}^{d}$ and variance $\bm{v} \in \mathbb{R}^{d}$ from $ \hat{\textbf{p}}$. These global features can be considered as high-level representations of the continuous 3D space, and they will be fed into the Prompt Learning Network, denote as $F_{PLN}$, to generate the scene-specific description.
%\vspace{-2pt}
\begin{equation}
	\label{eq5}
	\textbf p^{ss} = {F}_{PLN}\left(\left [\bm{\mu},\bm{v} \right ]\right),
\end{equation}
\noindent
where $\left[\cdot,\cdot \right ]$ denotes concatenation and $\textbf p^{ss}$ is scene-specific description. Note that, we stop the gradient in this branch to avoid backpropagation to the backbone. Therefore, given the corresponding $C_k= \left \{0,1,\cdots,n_k\right \}$ label set in this scene $ C \in \left \{1,2,\cdots,C\right \}$ Eq.~\eqref{eq1} was refactored as,% \in \mathbb{R}^{D}
\begin{equation}
\label{eq6}
	{\textbf T}_{k} ={F}_{text}\left(\left [\textbf p^{ss},c_{i} \right ]\right) , i \in \left \{1,2,\cdots,n_k\right \}.
\end{equation}
%In contrast to Eq.~\eqref{eq1}, 
Therefore, learnable prompts are able to generate more precise anchors in the latent space. In addition, the prompt allows the network to extract more domain-specific features rather than only relying on shared features in the same category. 
\section{Experiments}

\begin{table}
\centering
\fontsize{8}{14}\selectfont
\begin{tabular}{c|c|ccc}
% \begin{tabularx}{\linewidth}{>{\Centering}p{1.45cm}|>{\Centering}p{1.4cm}|>{\Centering}p{0.9cm}>{\Centering}p{0.9cm}>{\Centering}p{0.9cm}}
\hline
\multicolumn{1}{c|}{Backbone} &
\multicolumn{1}{c|}{Method}      & \multicolumn{1}{c}{OA \small{$\uparrow$}} & \multicolumn{1}{c}{mAcc \small{$\uparrow$}} & \multicolumn{1}{c}{mIoU \small{$\uparrow$}}\\ 
\hline
\multirow{5}{*}{\shortstack[l]{PointNet++~\citep{qi2017pointnet++}}} &N/A & 61.21                   & 47.42                    & 29.54             \\
&AD~\citep{ganin2015unsupervised}     & 64.48                  & 49.13                    & 31.54        \\
&Group DRO~\citep{sagawa2019distributionally}   & 63.39                  & 48.03                    & 29.99     \\
&V-REx~\citep{krueger2021out}                & 63.54                  & 48.04                    & 30.30     \\
% \rowcolor{LightCyan}
&\cellcolor{LightCyan}\textbf{Ours}             & \cellcolor{LightCyan}\textbf{69.03}   & \cellcolor{LightCyan}\textbf{51.44}      & \cellcolor{LightCyan}\textbf{36.79}   \\ \hline
\multirow{5}{*}{\shortstack[l]{ASSANet~\citep{qian2021assanet}} }  &N/A   & 70.25                  & 51.71                    & 35.29     \\
&AD~\citep{ganin2015unsupervised}       & 69.74                  & 50.86                    & 34.29       \\
&Group DRO~\citep{sagawa2019distributionally}      & 70.08                  & 52.42                    & 35.53     \\
&V-REx~\citep{krueger2021out}            & 70.79                  & 51.65                    & 35.17      \\
% \rowcolor{LightCyan}
&\cellcolor{LightCyan}\textbf{Ours}                              & \cellcolor{LightCyan}\textbf{71.09}         & \cellcolor{LightCyan}\textbf{56.28}           & \cellcolor{LightCyan}\textbf{40.42}     \\ \hline
\multirow{5}{*}{\shortstack[l]{PointNeXt-B~\citep{qian2022pointnext}}} &N/A   & 70.22                  & 49.00                    & 32.88             \\
&AD~\citep{ganin2015unsupervised}             & 70.01                  & 49.36                    & 33.37        \\
&Group DRO~\citep{sagawa2019distributionally}    & 70.48                  & 50.76                    & 34.26           \\
&V-REx~\citep{krueger2021out}                & \textbf{72.30}                  & 51.28                    & 35.50        \\
% \rowcolor{LightCyan}
&\cellcolor{LightCyan}\textbf{Ours}                              & \cellcolor{LightCyan}71.95         & \cellcolor{LightCyan}\textbf{55.25}           & \cellcolor{LightCyan}\textbf{40.50}      \\ 
\hline
\end{tabular}
\caption{Comparison of 2D DG/DA methods and our method in 3D-FRONT, S3DIS $\rightarrow$ ScanNet val (11 classes), best results in each backbone are mark in \textbf{bold}. In general, our method is more generalizable than other existing 2D methods.}
\label{tab: tab2}
\end{table}

\begin{table}
\centering
\fontsize{8}{14}\selectfont
\begin{tabular}{c|c|ccc}
% \begin{tabularx}{\linewidth}{>{\Centering}p{1.45cm}|>{\Centering}p{1.4cm}|>{\Centering}p{0.9cm}>{\Centering}p{0.9cm}>{\Centering}p{0.9cm}}
\hline
\multicolumn{1}{c|}{Backbone} &
\multicolumn{1}{c|}{Method}      & \multicolumn{1}{c}{OA \small{$\uparrow$}} & \multicolumn{1}{c}{mAcc \small{$\uparrow$}} & \multicolumn{1}{c}{mIoU \small{$\uparrow$}} \\ 
\hline
\multirow{5}{*}{\shortstack[l]{PointNet++~\citep{qi2017pointnet++}}} & N/A & 79.63                  & 56.91                    & 50.71                 \\
&AD~\citep{ganin2015unsupervised}                       & 71.64                  & 53.60                    & 40.27      \\
&Group DRO~\citep{sagawa2019distributionally}    & 66.8                  & 52.10                    & 43.44       \\
&V-REx~\citep{krueger2021out}    & 71.26                  & 52.48                    & 45.76             \\
% \rowcolor{LightCyan}
&\cellcolor{LightCyan}\textbf{Ours}                            & \cellcolor{LightCyan}\textbf{82.97}         & \cellcolor{LightCyan}\textbf{60.26}           & \cellcolor{LightCyan}\textbf{54.53}             \\ \hline
\multirow{5}{*}{\shortstack[l]{ASSANet~\citep{qian2021assanet}} }    & N/A & 80.19                  & 57.99                    & 49.33             \\
&AD~\citep{ganin2015unsupervised}                                & 79.88                  & 58.12                    & 49.64          \\
&Group DRO~\citep{sagawa2019distributionally}                         & 80.88                  & 58.19                    & 49.58        \\
&V-REx~\citep{krueger2021out}                              & 79.7                  & 58.52                    & 50.43          \\
% \rowcolor{LightCyan}
&\cellcolor{LightCyan}\textbf{Ours}                              & \cellcolor{LightCyan}\textbf{83.55}         & \cellcolor{LightCyan}\textbf{59.08}           &\cellcolor{LightCyan} \textbf{53.10}             \\ \hline
\multirow{5}{*}{\shortstack[l]{PointNeXt-B~\citep{qian2022pointnext}}}  & N/A & 79.96                  & 60.62                    & 53.26             \\
&AD~\citep{ganin2015unsupervised}          & 83.86                  & 61.61                    & 55.26               \\
&Group DRO~\citep{sagawa2019distributionally}     & 82.51                  & 61.08                    & 54.39     \\
&V-REx~\citep{krueger2021out}       & 82.63                  & 61.70                   & 54.75            \\
% \rowcolor{LightCyan}
&\cellcolor{LightCyan}\textbf{Ours}                              & \cellcolor{LightCyan}\textbf{87.51}         & \cellcolor{LightCyan}\textbf{64.59}           & \cellcolor{LightCyan}\textbf{58.80}             \\ 
\hline
\end{tabular}
\caption{Comparison of 2D DG/DA methods and our method in 3D-FRONT, ScanNet $\rightarrow$ S3DIS-Area 5 (11 classes), \textbf{bold} denotes best performance in corresponding backbone. Our method steadily overwhelms other methods under this task.}
\label{tab: tab3}
\end{table}

\paragraph{Datasets.}
\textit{S3DIS}~\citep{armeni20163d} is a real-world indoor 3D point cloud dataset, which is widely used for the performance evaluation of 3D semantic segmentation. There are 271 rooms across 6 areas and be annotated as 13 categories at point level. Following previous works~\citep{qi2017pointnet++,zhao2021point}, We choose area-5 as the validation set. Other areas are all used for training. \textit{ScanNetV2}~\citep{dai2017scannet} is a large-scale real-world indoor 3D point cloud dataset. it contains 20 different categories in 1201 scenes for training, 312 scenes for validation, and 100 scenes for testing. \textit{3D-FRONT}~\citep{fu20213d} is a synthesis indoor 3D mesh dataset that covers 18968 rooms from 3D-FUTURE, which span 31 scene categories and 34 object semantic super-classes. In our experiments, we use a sub-point cloud dataset processed by~\citep{ding2022doda}, which contains 4995 rooms for training.

\paragraph{Network architecture.} We verify our approach with three 3D segmentation models, PointNet++~\citep{qi2017pointnet++}, ASSANet~\citep{qian2021assanet}, PointNeXt~\citep{qian2022pointnext}, which have been widely used as baselines for 3D point cloud segmentation tasks. For the text encoder, we choose CLIP which provides powerful semantic latent space and has been widely proven in zero-shot perception tasks~\citep{zhang2022pointclip,li2022language}. We further analyzed various text encoders of pre-trained language models and the results are shown in Sec.~\ref{TextEncoders}.

\paragraph{Implementation detail.} We follow the training strategy and data augmentation in~\citep{qian2022pointnext}. To be more specific, we train our framework using AdamW~\citep{loshchilov2017decoupled} optimizer and set initial learning rate as $lr=0.001$ and gradually decays at [70, 90] epochs with a decay rate of 0.1 and we train our framework for 100 epochs.
%from here
For the evaluation on Domain Generalization, we train our data with batch size 3 in each source and using 32000 points per batch and assume target domain is inaccessible. For the evaluation on Transfer Learning we train our data from 3 given heterogeneous sources with batch size 3 in each source and using 24000 points per batch, and select the best model on validation for testing. For fair comparisons, all results are reported with the same training strategy and data augmentation. 
% \subsection{Experiments setting}
\subsection{Evaluation on Domain Generalization}

We consider domain generalization, which is a challenging situation in dynamic environments to investigate the generalization ability of our framework. We assume that data distributions are unknown until inference. To train traditional methods, we manually condense labels from different sources follow~\citep{ding2022doda} and consider such labels as objects of interest to predict in the target domain. We then reproduce three popular Domain Generalization or Domain Adaptation methods, i.e., AD~\citep{ganin2015unsupervised}, Group DRO~\citep{sagawa2019distributionally} and V-REx~\citep{krueger2021out}. 
Our implementation details are explained in the supplementary material.
% \subsection{Result}
% \begin{figure*}[!htbp]
%  \centering
%    \includegraphics[width=\textwidth]{figures/fig_5.pdf}
%    % \vspace{-13.5em} %-13.5em
% \caption{Visualization of the network prediction. (a). the examples to show synonyms labels have the same result. (b).the examples to show our model work well with hierarchical labels of different granularity. View from left to right, the replaced labels are \underline{underlined} and both the replacement and replaced labels are mark in \textbf{bold}.}
% \label{fig:exp_qualitative}
% \end{figure*}

\subsubsection{Comparing with the State-of-the-art}
As shown in Tab.~\ref{tab: tab2} and Tab.~\ref{tab: tab3}, our model  outperforms other approaches by a large margin, \emph{i.e.} On the experiments of 3D-FRONT, S3DIS$\rightarrow$ ScanNet, our method imporve \textbf{7.25$\%$} mIoU in PointNet++, \textbf{5.13$\%$} mIoU in ASSANet and \textbf{7.62$\%$} mIoU in PointNeXt-B. \emph{i.e.} On the experiments of 3D-FRONT, ScanNet$\rightarrow$ S3DIS. Our method imporve \textbf{3.82$\%$} mIoU over PointNet++, \textbf{3.77$\%$} mIoU over ASSANet, \textbf{5.54$\%$} mIoU over PointNeXt-B. Especially, our model is particularly favored in the small model (PointNet++) and small dataset (S3DIS), which are more challenging. 
Besides, it should be noted that only 11 labels can be well aligned, and thus the evaluation is done only on these labels. The baseline models cannot handle other labels as they do not appear during training. In contrast, benefiting from the shared label space,  our model can make a reasonably good prediction, as will be shown in Fig. \ref{fig:exp_qualitative}.  
Meanwhile, we also observe that when training on the unified dataset with large data shift (3D-FRONT, ScanNet$\rightarrow$ S3DIS), existing DG and DA approaches do not generalize well, and sometimes even hinder the model performance. In contrast, our method robustly improve generalization of each backbone.
\begin{table}[!t]
\centering
\fontsize{7.5}{14}\selectfont
\begin{tabular}{l|c|lll}
\hline 
\multicolumn{1}{c|}{Text Encoder} & \multicolumn{1}{c|}{\# Dimension} & \multicolumn{1}{c}{OA\small{$\uparrow$}} & mAcc\small{$\uparrow$}  & mIoU\small{$\uparrow$} \\ \hline
BERT~\citep{devlin2018bert}  & 768                               & 81.72                   & 57.68 & 44.92    \\
ALBERT~\citep{lan2019albert}     & 768                               &66.92               &53.77 & 42.80  \\
T5~\citep{raffel2020exploring}     & 768                               &86.49                   &64.21  &57.13   \\
ViT/base-32(CLIP)~\citep{radford2021learning}    & 512                               &87.65                 &64.50 & 58.14 \\
RN50X4(CLIP)~\citep{radford2021learning}  & 640                               &87.05                   & 63.35    &57.62  \\
RN50X16(CLIP)~\citep{radford2021learning}   & 768                               &\textbf{87.51}                   &\textbf{64.59}   &\textbf{58.80}  \\ 
\hline 
\end{tabular}

\caption{Analysis on various text encoders. We evaluate on 3D-FRONT, ScanNet$\rightarrow$ S3DIS task. Best results are mark in \textbf{bold}.}
\label{tab: text encoder}
\end{table}

\begin{table}[]
\centering
\small
%\centering
\fontsize{8}{15}\selectfont
\setlength{\tabcolsep}{4mm}{
\begin{tabular}{c|cccc}
\hline
\# Token Length & 0                    & 4                    & 8                    & 16                   \\ \hline
mIoU\small{$\uparrow$}   & 57.00                & 57.66                & \textbf{58.80}       & 57.68                \\
$\Delta$       & \multicolumn{1}{c}{-} & \textcolor{green}{+0.66} &\textcolor{green}{+1.80} &\textcolor{green}{+0.68}\\ \hline
\end{tabular}}
\caption{Analysis on learnable token length. We evaluate on 3D-FRONT, ScanNet$\rightarrow$ S3DIS task. 0 refers to MantraNet without prompt learning network. The \textcolor{green}{green} indicates relative gains and \textbf{bold} indicates best result.}
\label{tokenlength}
\end{table}

\subsubsection{Analysis}
In this section, we investigate our proposed MantraNet through extensive studies. All experiment are conducted under the same evaluation settings in 3D-FRONT, and ScanNet $\rightarrow$ S3DIS.
% \hspace*{\fill} \\
\paragraph{Text Encoders.}\label{TextEncoders} In principle, various language pre-trained models can be used for discovering semantic anchor of labels. We test several encoders that train in pure texts, such as BERT~\citep{devlin2018bert}, ALBERT~\citep{lan2019albert} and T5~\citep{raffel2020exploring}, and those trained with text-image pairs~\citep{radford2021learning} in Tab.~\ref{tab: text encoder}. We find that the text encoders which train in text-image pairs consistently outperform those that train in pure texts, and RN50X16 provide by CLIP achieves the best performance and outperforms T5 that trains in pure text by $1.67\%$. Such result shows that CLIP maintains a better latent space that aligns image appearance and text representation than those trained only with pure text. Therefore, we adopt CLIP as our text encoder, and it can distinguish label semantic representation more accurately and increases the overall generalization ability of our MantraNet. 
% \hspace*{\fill} \\

\paragraph{Learnable Context Length.} In Tab.~\ref{tokenlength}, We conduct the ablation study about learnable tokens length. Follow~\citep{zhou2022learning}, we study 0, 4, 8, 16 learnable context tokens, 0 means the framework without a prompt learning network. We observe that 8 tokens achieved the best result, which indicate that if we have fewer learnable prompts, it does not contain enough information to smooth the covariance shift. However, if we add too many tokens, we will introduce ambiguities that interfere with the semantic representation of the label itself.

% \subsection{Practical Application}
\subsection{Evaluation on Transfer Learning}
%target small dataset, argue that our methods can levearge more dataset that outperforms other methods. So the settings is we use 2 dataset and inference on 1.
In this section, we evaluate our model in a more practical setting. In practice, when faced with a small sample size problem, the deep neural network may easily overfit the training data, which lead to inferior performance. In this sense, one may want to leverage all available data on hand to regularize the network, so that the model performs well on the small target dataset.

% In practice, we often have limited training data in small dataset, therefore, traditional methods suffer under such condition. However, our ManchaNet allows unified training with heterogeneous data sources, and still generalize well when training data are limited. 
% To investigate how our method performs under such extreme condition, we choose the task in S3DIS-Area5, which only have 203 training data. We then combine 3 datasets together, which are ScanNetV2~\citep{dai2017scannet},S3DIS~\citep{armeni20163d} and 3D-FRONT~\citep{fu20213d}, and inference on S3DIS~\citep{armeni20163d}. For fair comparison, we also adopt the same setting to traditional methods by adjusting their decoder architecture. We report our result in Tab~\ref{tab: tab1}, and our ManchaNet outperforms all baselines with a large margin under such extreme settings.
To investigate how our method performs in this case, we take the S3DIS~\citep{armeni20163d} as the target dataset and use ScanNetV2~\citep{dai2017scannet} and 3D-FRONT~\citep{fu20213d} to augment the training data. Note that S3DIS is a small dataset, i.e., it contains only 203 training data. ScanNetV2 is a medium-scale realistic dataset, and 3D-FRONT is a big synthetic one. This simulates the situation commonly appearing in practice. Also, note that the label sets are not well-aligned, and most existing methods are not applicable. We thus compare our model with two baselines, i.e., Target-only and Multi-Head. Target-only is trained with the only target domain. Multi-Head uses 3 different prediction heads for the three datasets. As shown in Tab. \ref{tab: tab1}, our model significantly outperforms the two baselines, \emph{i.e.} , our method yields \textbf{2.59$\%$} and \textbf{5.01$\%$} mIoU improvement over the Target-only and Multi-Head in PointNet++ and \textbf{0.86$\%$} and \textbf{2.22$\%$} mIoU improvement in ASSANet. Meanwhile, In PointNeXt-B, \textbf{1.41$\%$} and \textbf{1.92$\%$} mIoU gains are obtained in our method.

\begin{table}[]
\centering
%%%TRY$\uparrow$}
%\scriptsize
\small
%\resizebox{\linewidth}{!}{
\fontsize{8}{15}\selectfont
\begin{tabular}{c|c|ccc}
\hline
\multirow{2}{*}{Backbone}    & \multirow{2}{*}{Method} & \multicolumn{3}{c}{S3DIS Area-5}                  \\ \cline{3-5} 
                             &                         & OA\small{$\uparrow$}     & mAcc\small{$\uparrow$}  & mIoU  \small{$\uparrow$}  \\ \hline
\multirow{3}{*}{PointNet++~\citep{qi2017pointnet++}}  &T & 88.17          & 70.79          & 63.99          \\
                             &MH         & 87.46          & 68.54          & 61.57          \\
                             &Ours       & \textbf{88.88} & \textbf{73.46} & \textbf{66.58} \\ \hline
\multirow{3}{*}{ASSANet~\citep{qian2021assanet}}     & T                       & 88.86          & 72.36          & 66.07          \\
                             &MH         & 88.73          & 71.09          & 64.71          \\
                             &Ours         & \textbf{89.47} & \textbf{73.49} & \textbf{66.93} \\ \hline
\multirow{3}{*}{PointNeXt-B~\citep{qian2022pointnext}} & T                       & 89.46          & 74.22          & 67.83          \\
                             &MH      & 89.37          & 73.79          & 67.32          \\
                             &Ours       & \textbf{90.38} & \textbf{75.12} & \textbf{69.24} \\ \hline
\end{tabular}%}
\caption{Quantitative results on S3DIS Area-5. T (Target-only) means traditional method only training on data from the target domain, MH (Multi-head) means traditional method training on all datasets by setting multiple task specific head. The \textbf{bold} denotes the best performance of each backbone.}
\label{tab: tab1}
\end{table}
% \subsection{Result}
\begin{figure*}[!htbp]
 \centering
   \includegraphics[width=0.95\linewidth]{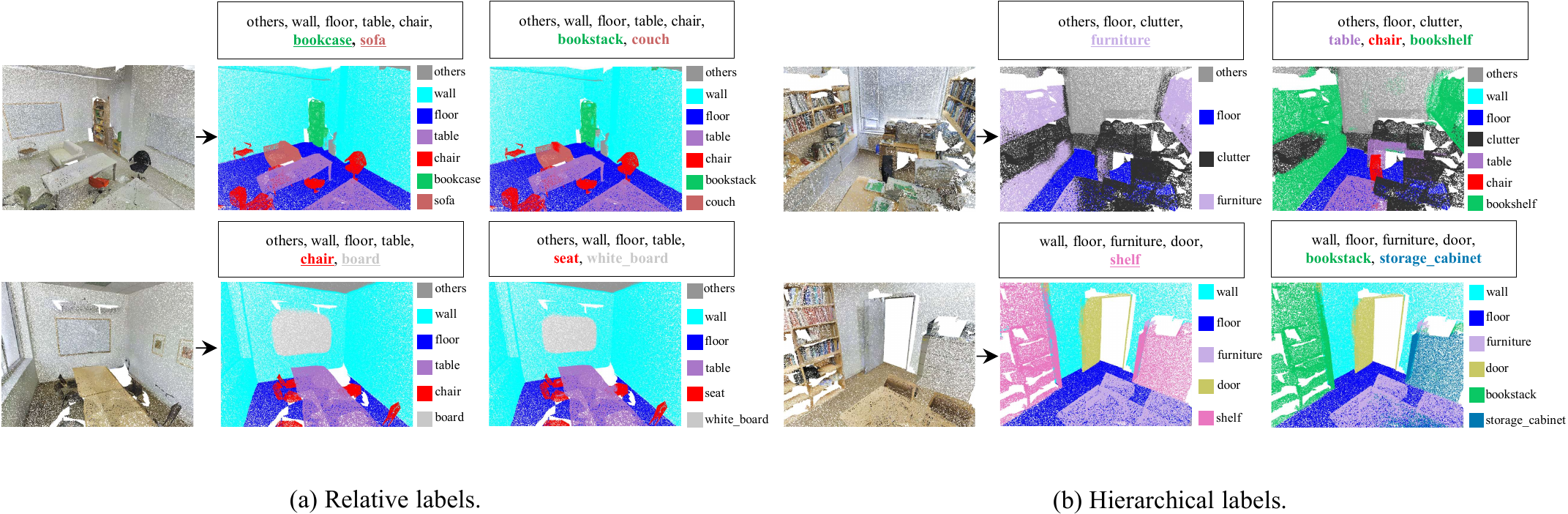}
   % \vspace{-13.5em} %-13.5em
\caption{Visualization of the network prediction. (a). the examples to show synonyms labels have the same result. (b). the examples to show our model work well with hierarchical labels of different granularity. View from left to right, the replaced labels are \underline{underlined} and both the replacement and replaced labels are mark in \textbf{bold}.}
\label{fig:exp_qualitative}
\end{figure*}
\subsubsection{Qualitative Evaluation}\label{qf}
In this section, we provide a case study to better understand our proposed framework. To obtain a model with better generalization and segmentation capabilities for practical requirements, we unify training set of S3DIS, SCANNET and 3D-FRONT for training and apply this model in the following qualitative evaluation.
% \hspace*{\fill} \\
% \noindent 
\paragraph{Related but previously unseen labels.} Fig.~\ref{fig:exp_qualitative}(a) visualize the segmentation maps for given related but previously unseen labels. In the first row, we set the required label as "\textit{others}", "\textit{wall}", "\textit{floor}", "\textit{table}", "\textit{chair}", "\textit{bookcase}" and "\textit{sofa}", and the results show that the model is able to classify such points that belong to the required labels. When we replace the given label "\textit{bookcase}" and "\textit{sofa}" to "\textit{bookstack}" and "\textit{couch}", which have similar semantic meanings but are not present in the training label sets. MantraNet returns a similar segmentation map as before, which shows that our method is able to generalize to unseen labels belonging to the same category. In the second row, MantraNet correctly assigns the unseen label "\textit{seat}" and "\textit{white board}" to the points that correspond to "\textit{chair}" and "\textit{board}". This result shows that MantraNet is good to extend perceptual capabilities to new categories through shared space.
% \hspace*{\fill} \\
% \noindent 
\paragraph{Hierarchical unseen labels.} Fig.~\ref{fig:exp_qualitative}(b) illustrates that MantraNet can handle coarse label to fine label segmentation which is often found in a variety of practical requirements. In the first row, we start with "\textit{others}", "\textit{floor}", "\textit{clutter}" and highly generalized labels "\textit{furniture}" for segmentation, please note that "\textit{others}" and "\textit{furniture}" does not appear in the training label sets. MantraNet successfully recognizes the coarse label and it can predict the fine labels when requested, such as ("\textit{table}", "\textit{chair}", "\textit{bookshelf}"). In addition, we show a salient case about splitting a seen category into two unseen categories in the second row, "\textit{storage carbinet}" is also a new label for our MantraNet. Because of the advantages of using a well-trained shared space, MantraNet can still perceive the points belonging to this unseen category.

\section{Concluding Remarks}

% This paper presents the MantraNet, a principled solution that supports joint training with multiple datasets of different label sets and data distribution. This addresses a critical problem in point cloud segmentation neglected for a long time, i.e., the isolated island of data. 
"\textit{The isolated island of data}" is a critical barrier that hinders the practical usability of modern point cloud segmentation approaches. Point cloud segmentation datasets are different in data distribution and label sets. Therefore, existing methods cannot access a significant amount of data to train a generalizable model that can fit complicated practical needs. In this work, we propose the MantraNet, a principled solution that supports joint training with multiple datasets of different label sets and data distribution. As such, our model suppresses the state-of-the-art methods in different benchmarks under practical settings by a large margin. 

It should be noted that this idea is not limited to point cloud segmentation. In many real-world applications, different datasets are collected from different sources for different purposes, and thus they are heterogeneous by nature. By leveraging label names or other textual side information, one may conduct unified training with all of them. This allows for fully leveraging all available data to learn more shareable knowledge. This will be explored in our future work. 
% Different datasets have different distribution and label sets, which makes it challenging for

\clearpage
% {\small
% \bibliographystyle{ieee_fullname}
% \bibliography{egbib}
% }

\bibliography{egbib}
% \clearpage

% \iffalse
% \appendix
% \input{LaTeX/chapter/appendix}
% \clearpage

\end{document}